\documentclass[sigconf]{acmart}
 \pdfoutput=1
 
\renewcommand\footnotetextcopyrightpermission[1]{} 
\usepackage{balance}

\def\BibTeX{{\rm B\kern-.05em{\sc i\kern-.025em b}\kern-.08emT\kern-.1667em\lower.7ex\hbox{E}\kern-.125emX}}
    
\copyrightyear{2019}
\acmYear{2019}
\setcopyright{acmcopyright}
\acmConference[KDD '19] {The 25th ACM SIGKDD Conference on Knowledge Discovery and Data Mining}{August 4--8, 2019}{Anchorage, AK, USA}
\acmPrice{15.00}
\acmDOI{10.1145/3292500.3330694}
\acmISBN{978-1-4503-6201-6/19/08}

\begin{document}
\title{OCC: A Smart Reply System for Efficient In-App Communications}

\author{Yue Weng}
\authornote{Equal Contribution}
\affiliation{%
  \institution{Uber AI}
  \city{San Francisco}
  \state{California}
  \postcode{94103}
}
\email{yweng@uber.com}

\author{Huaixiu Zheng}
\authornote{Equal Contribution}
\affiliation{%
  \institution{Uber AI}
  \city{San Francisco}
  \state{California}
  \postcode{94103}
}
\email{huaixiu.zheng@uber.com}

\author{Franziska Bell}
\affiliation{%
  \institution{Uber AI}
  \city{San Francisco}
  \state{California}
  \postcode{94103}
}
\email{fran@uber.com}

\author{Gokhan Tur}
\affiliation{%
  \institution{Uber AI}
  \city{San Francisco}
  \state{California}
  \postcode{94103}
}
\email{gokhan@uber.com}

\begin{abstract}

Smart reply systems have been developed for various messaging platforms. In this paper, we introduce Uber's smart reply system: one-click-chat (OCC), which is a key enhanced feature on top of the Uber in-app chat system. It enables driver-partners to quickly respond to rider messages using smart replies. The smart replies are dynamically selected according to conversation content using machine learning algorithms. Our system consists of two major components: intent detection and reply retrieval, which are very different from standard smart reply systems where the task is to directly predict a reply. 
It is designed specifically for mobile applications with short and non-canonical messages.
Reply retrieval utilizes pairings between intent and reply based on their popularity in chat messages as derived from historical data. For intent detection, a set of embedding and classification techniques are experimented with, and we choose to deploy a solution using unsupervised distributed embedding and nearest-neighbor classifier. It has the advantage of only requiring a small amount of labeled training data, simplicity in developing and deploying to production, and fast inference during serving and hence highly scalable. At the same time, it performs comparably with deep learning architectures such as word-level convolutional neural network. Overall, the system achieves a high accuracy of $76\%$ on intent detection. Currently, the system is deployed in production for English-speaking countries and $71\%$ of in-app communications between riders and driver-partners adopted the smart replies to speedup the communication process.

\end{abstract}

%


\maketitle
\section{Introduction} \label{intro}

\begin{figure}
\includegraphics[width=0.2\textwidth]{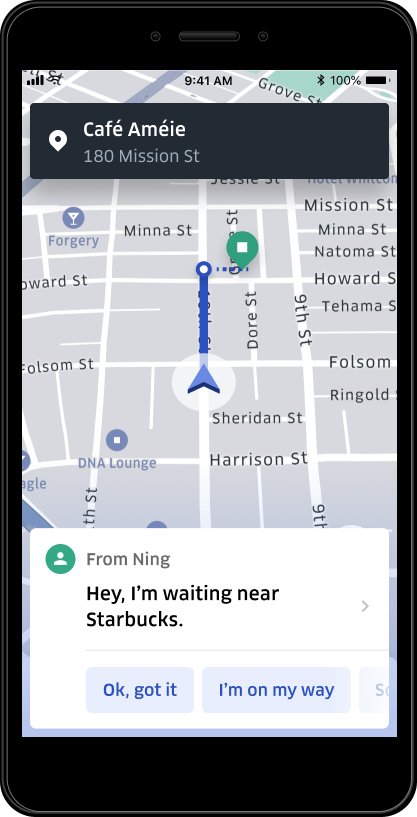}
\caption{With one-click chat, driver-partners can more easily respond to rider messages.}
\label{fig:occ_demo}
\end{figure}

Uber's ride-sharing business connects driver partners to riders who need to transport around cities. The app provides navigation and many other innovative technology and features that assist driver-partners with finding riders at the pick-up locations. 
Uber's in-app chat \cite{uberchat}, a real-time in-app messaging platform launched in early 2017, is one of them. 
Before having the functionality to chat within the app, communication between customers occurred outside of the mobile app experience using third-party technologies. 
This resulted in safety concerns, higher operational costs, fewer completed trips, and most importantly, limits the company's ability to understand and resolve challenges both riders and driver partners were having while using the app. Although the newly added chat feature has solved many of these problems by bringing the chat experience into the app, it still requires driver-partners to type messages while driving, which is a huge safety concern. 
According to a public study, compared to regular driving, accident risk is about 2.2 times higher when talking on a hand-held cell phone and 6.1 times higher when texting \cite{phoneinteract}. 
Therefore, to provide a safe and smooth in-app chat experience for driver-partners, we developed One-Click Chat (OCC), a smart reply system that allows driver-partners to respond to messages using smart replies selected dynamically according to the conversation context, as shown in Figure~\ref{fig:occ_demo}.

There has been a surge of interest in developing and using smart reply and chatbot systems on commercial platforms \cite{DBLP:journals/corr/KannanKRKTMCLGY16, DBLP:journals/corr/HendersonASSLGK17, DBLP:conf/chi/XuLGSA17}. 
However, building an intelligent system to automatically generate suggested replies is not a standard machine learning problem. Anjuli et al. \cite{DBLP:journals/corr/KannanKRKTMCLGY16} proposed to divide the task into two components: predicting responses and identifying a target response. Specifically, to predict responses, they leveraged a sequence-to-sequence (Seq2seq) \cite{DBLP:conf/nips/SutskeverVL14} framework with long short-term memory (LSTM) \cite{lstm} trained on large-scale email conversations; to obtain the final response, they first proposed a semi-supervised approach to generate a response pool and then select from it based on the LSTM predictions to control the actual replies as free text generation is still not mature enough for commercial use \cite{DBLP:journals/corr/KannanKRKTMCLGY16}. Similarly, LinkedIn used a statistical model for predicting responses for incoming messages \cite{linkedin}.

In contrast, instead of predicting responses directly, our work experiments with techniques that perform language understanding (i.e., intent detection) and reply retrieval separately. Our approach requires a much smaller scale labeled dataset for training. Compared to the generic smart replies, OCC is designed for Uber's domain-specific use case to streamline communications between driver-partners and riders during the pick-up stage.
In addition, in-app messages on the Uber platform are typically very short (averaging $4$-$5$ words) and non-canonical (with typos, abbreviations etc.) compared to other platforms such as email. This poses unique challenges to designing and developing such a smart reply system.
In this paper, we share our experiences building and integrating Uber's smart reply system. The main contributions of this work
are as follows:
\begin{itemize}
\item Introducing an end-to-end smart reply system architecture, a mobile-friendly solution, in Section \ref{occ}. 
\item
Presenting a novel approach to break the task of smart reply down into two steps - intent detection and reply retrieval, tailored specifically for short messages on the mobile platform.
\item Proposing a mixture of unsupervised embedding and nearest-neighbor supervised learning approaches which do not require a large amount of labeled data. This combined approach achieves comparable performance to deep learning architectures, but is much easier to implement and deploy. The step-by-step algorithmic approach is discussed in Section \ref{MLA}.
\item Conducting comprehensive experiments to compare different models and uncover their underlying mechanisms and
shortcomings. Experiments are discussed in Section \ref{experiment}.
\end{itemize}

\begin{figure}
\includegraphics[width=0.5\textwidth]{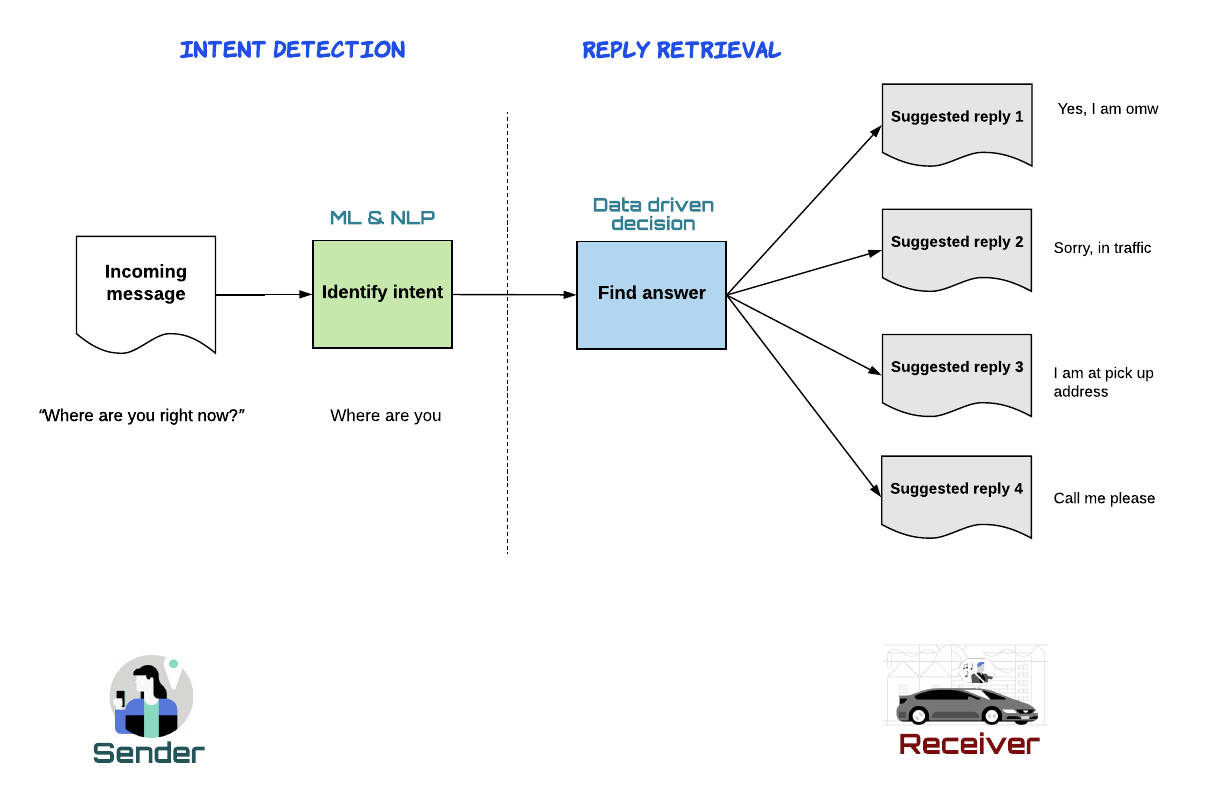}
\caption{\label{fig:zoomin}The machine learning algorithm empowers the flow of the OCC experience. Two key steps are involved: 1) intent detection and 2) reply retrieval.
}
\label{figure:intent_reply}
\end{figure}

\begin{figure}
\includegraphics[width=0.45\textwidth]{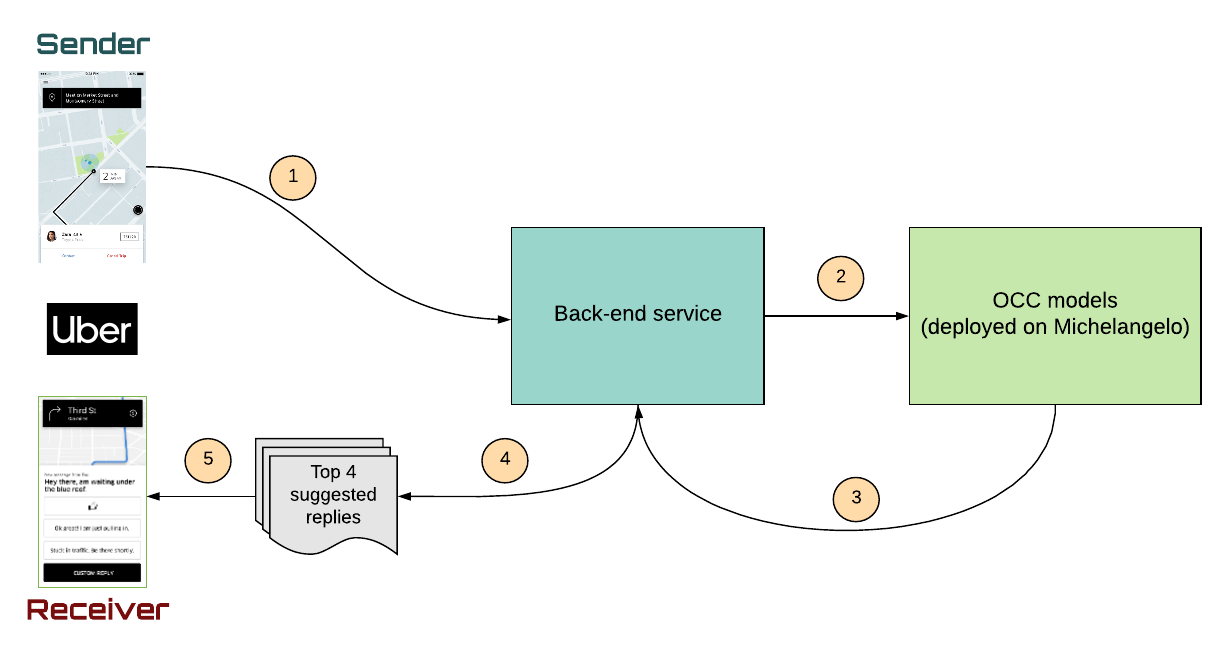}
\caption{The architecture for Uber's smart reply system, OCC, consists of a five-step workflow.}
\label{fig:occ_workflow}
\end{figure}

\section{One-Click Chat System} \label{occ}

As one of the world's largest and most recognized rider-sharing providers, there are hundreds and thousands of messages exchanged on the platform every day. OCC, one of the latest key enhanced features on our chat platform, aims to provide driver-partners with a one-click chatting experience by offering them the most relevant replies. 

To find the best replies to each incoming message, we formulate the task into a machine learning problem with two major components:

\begin{enumerate}
\item Intent Detection
\item Reply Retrieval
\end{enumerate}

Figure ~\ref{figure:intent_reply} illustrates how OCC works in the real world.
Specifically, a driver-partner receives an incoming rider message asking \textit{Where are you right now?}, which is very common during pick-up. In intent detection, the OCC system detects the intent of the message as \textit{Where are you?}. Then in reply retrieval, the system surfaces the top four most relevant replies to the driver-partner, which, in this example, are \textit{Yes, I am omw}, \textit{Sorry, in traffic}, \textit{I am at pick-up address}, and \textit{Call me please}. Now, the driver-partner can select one of these four replies and send it back to the rider with a single tap. The above process finishes one round of communication with smart replies.


OCC system is fully integrated with our in-house chat platform.
As depicted in Figure \ref{fig:occ_workflow}, the system architecture follows a five-step workflow:

\begin{enumerate}
\item Sender (rider app) sends a message to driver partner.
\item Mobile side triggers and sends the message to a back-end service that calls the machine learning model hosted on Uber's in-house machine learning platform Michelangelo \cite{michelangelo}.
\item The model preprocesses and encodes the message, generates prediction scores for each possible intent, and sends them back to the back-end service.
\item Once the back-end service receives the predictions, it follows a predefined reply retrieval policy to find the best replies (in this case, the top four).
\item Receiver (driver-partner app) receives the smart replies and renders them for the driver-partner to select.
\end{enumerate}

\section{Machine Learning Methodology} \label{MLA}
\begin{figure}
\includegraphics[width=0.4\textwidth]{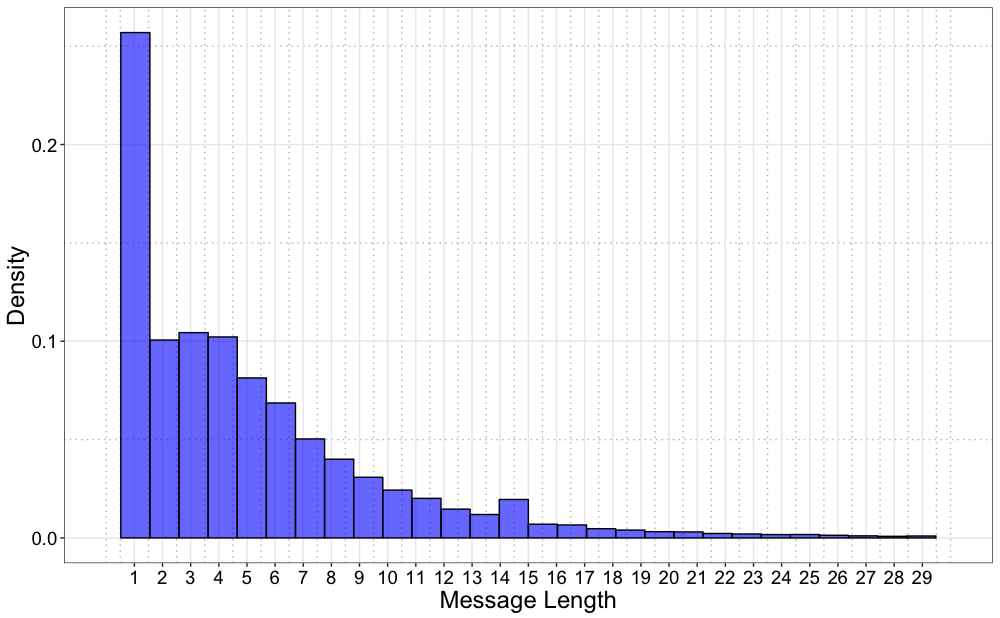}
\caption{\label{fig:message_freq}Chat message length frequency, on average $4$-$5$ words per message.
}
\label{figure:message_len}
\end{figure}

By design, OCC aims to provide an easy chat experience for driver-partners during the pick-up stage for Uber-specific scenarios and topic domains.
As a result, it shares a few technical challenges that are unique to mobile messaging systems:

\begin{itemize}
\item Messages are short compared to email or other communication channels, $4-5$ words per message on average given it is mostly used during pick-up, see Figure~\ref{figure:message_len} for the message length statistics. 
\item Messages are non-canonical, containing abbreviations, typos, and colloquialisms. 
Even for simple message like \textit{Where are you}, there are many variations including \textit{where r u :) ?}, \textit{w Here are you} and more. 
\end{itemize}


We designed our machine learning system with these challenges in mind, and adopted a mixture of unsupervised embedding and supervised classification techniques to tackle them accordingly. This section describes each component of the pipeline shown in Figure 2 in detail. 

\subsection{Features and Data}
We used millions of encrypted and anonymized historical in-app conversation data for our unsupervised embedding model. For supervised classification model, we collected and annotated thousands of conversational messages. Each of which is labeled as one of the intents in our system (such as \textit{I am here}). Here, we assume the messages are all single intent and validate the assumption manually. For the majority of the messages, they are short and convey a single intent in a single exchange of communication, even though there are exceptions. For this version of OCC, we use the text message as the only feature in the modeling process. For future iterations, contextual features such as length of conversation and trip information may be leveraged by the models.

\subsection{Intent Detection}
Given the nature of our message data (short and non-canonical with typos, etc.), we decide to put the emphasis on intent detection.
As we tackle intent detection \cite{DBLP:conf/wsdm/BrenesGP09}, we encounter several technical challenges due to the complexity of human language itself and the nature of messages exchanged on a mobile platform. For instance, there are many ways to ask the same question, such as \textit{Where are you going?}, \textit{Where are you heading?}, and \textit{What's your destination?}. With typos and abbreviations, chat messages introduce even more permutations. 
In addition, chat messages are typically very short, which makes distinguishing them from each other very challenging. 
Creating a system with replies for millions of individual questions does not scale, so we need a system that can identify the intent or topic behind each question, allowing us to provide replies to a finite set of intents. 

We formulate the language understanding task as a classification problem in order to have full control over message replies. We experimented with four different approaches for intent detection.  

\begin{itemize}
\item Frequency-based, a context-agnostic approach that suggests intents based on their frequency. 
\item CNN-based deep learning approach, both word and character level \cite{DBLP:conf/emnlp/Kim14}.
\item Embedding \cite{DBLP:journals/corr/LeM14} plus nearest neighbour classifier (NNC), which is a combination of unsupervised and supervised learning approach. It requires much smaller labeled data and performs on par with deep learning methods on a test dataset. 
\end{itemize}

Since both frequency and CNN-based approaches are relatively straight forward, we focus on the embedding-based NNC for the rest of the section.

\subsubsection{Message Embeddings}

\begin{figure}
\includegraphics[width=0.5\textwidth]{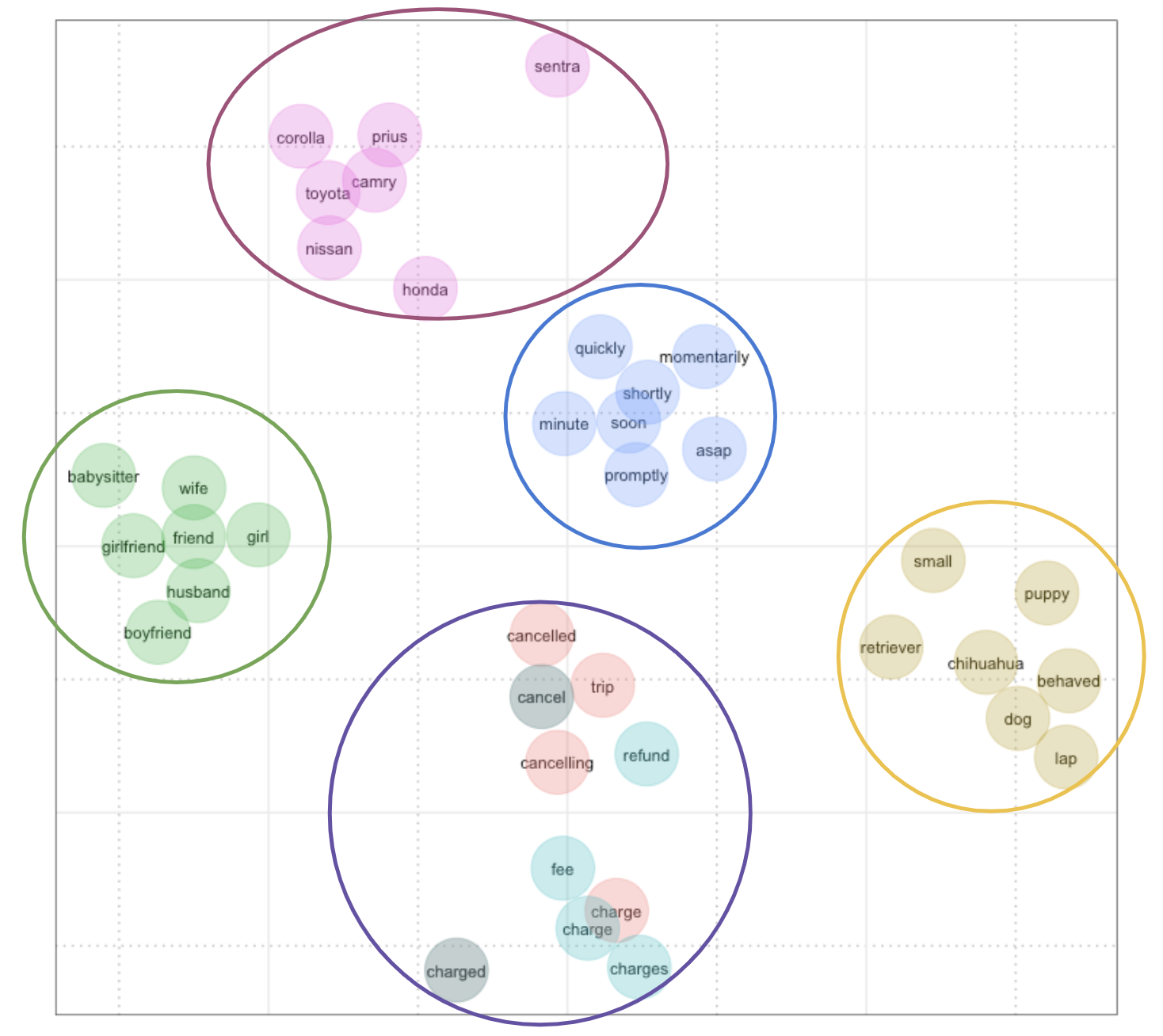}
\caption{\label{fig:architecture}This two-dimensional t-SNE projection of the Doc2vec word embedding illustrates the ability of the model to automatically organize concepts and learn implicitly the relationships between words, clustering them based on semantics.}
\label{figure:word_embedding}
\end{figure}

We embedded messages using the Doc2vec model \cite{DBLP:journals/corr/LeM14}, an unsupervised algorithm proposed by Le and Mikolov (2014), that learns fixed-length feature representations from variable-length pieces of text, such as sentences, paragraphs, and documents. Because our messages are domain-specific and contain a lot of typos and abbreviations, we decided to train our own embedding model using in-house data. Our Doc2vec model was trained on millions of anonymized, aggregated in-app chat messages and was then used to map each message to a dense vector embedding space.
Figure ~\ref{figure:word_embedding} visualizes the word vectors in a two-dimensional projection using a t-SNE plot \cite{Maaten2008VisualizingDU}. Since it captures the semantic meaning of words, the model can cluster similar words together. For example, \textit{fee} is close to \textit{charge} and \textit{refund}, but far away from \textit{friend}.

More formally, given a sequence of training words $w_1, w_2, .... , w_T$ from the document $D$, the objective of the Doc2vec model is to use a neural network to find the parameter sets $\theta^*$ in order to maximize the conditional probability of a target word $w_t$ given $k$ contextual words before and after the target word,
\begin{equation} \label{eq:1}
\theta^* = \text{argmax}_{\theta}\prod_{t\in S}P(w_t | w_{t-k}, ..., w_{t+k}; \theta)
\end{equation}

\noindent where $S$ is a sample of words from $D$ and $\theta\in {\theta_{d}, \theta_{w}, \theta_{\text{softmax}}}$, where $\theta_w$ are word vectors, $\theta_{\text{softmax}}$ contains the weights for a softmax hidden layer and $\theta_d$ are paragraph vectors. In short, the algorithm itself has two stages: 1) \textit{training stage} to optimize word vectors $\theta_w$, softmax weights $\theta_{\text{softmax}}$ and paragraph vectors $\theta_d$ on already seen paragraphs to maximize the probability of target word given contextual words; and 2) \textit{the inference stage} to compute paragraph vectors $d$ for new documents, which can be never seen before, by gradient descending on $d$ while holding $\theta_{\text{softmax}}$ and $\theta_w$ fixed \cite{DBLP:journals/corr/LeM14}.


\subsubsection{NNC Approach}
We build a nearest-neighbor classifier on top of the distributed representation of labeled messages from document embedding model Doc2vec.
The main motivation is that we have a relatively small set of (thousands of) labeled data to train a classification model. Overfitting is a big concern, and hence the non-parametric nearest-neighbor classifier can largely avoid such a problem.

\begin{figure}
\includegraphics[width=0.35\textwidth]{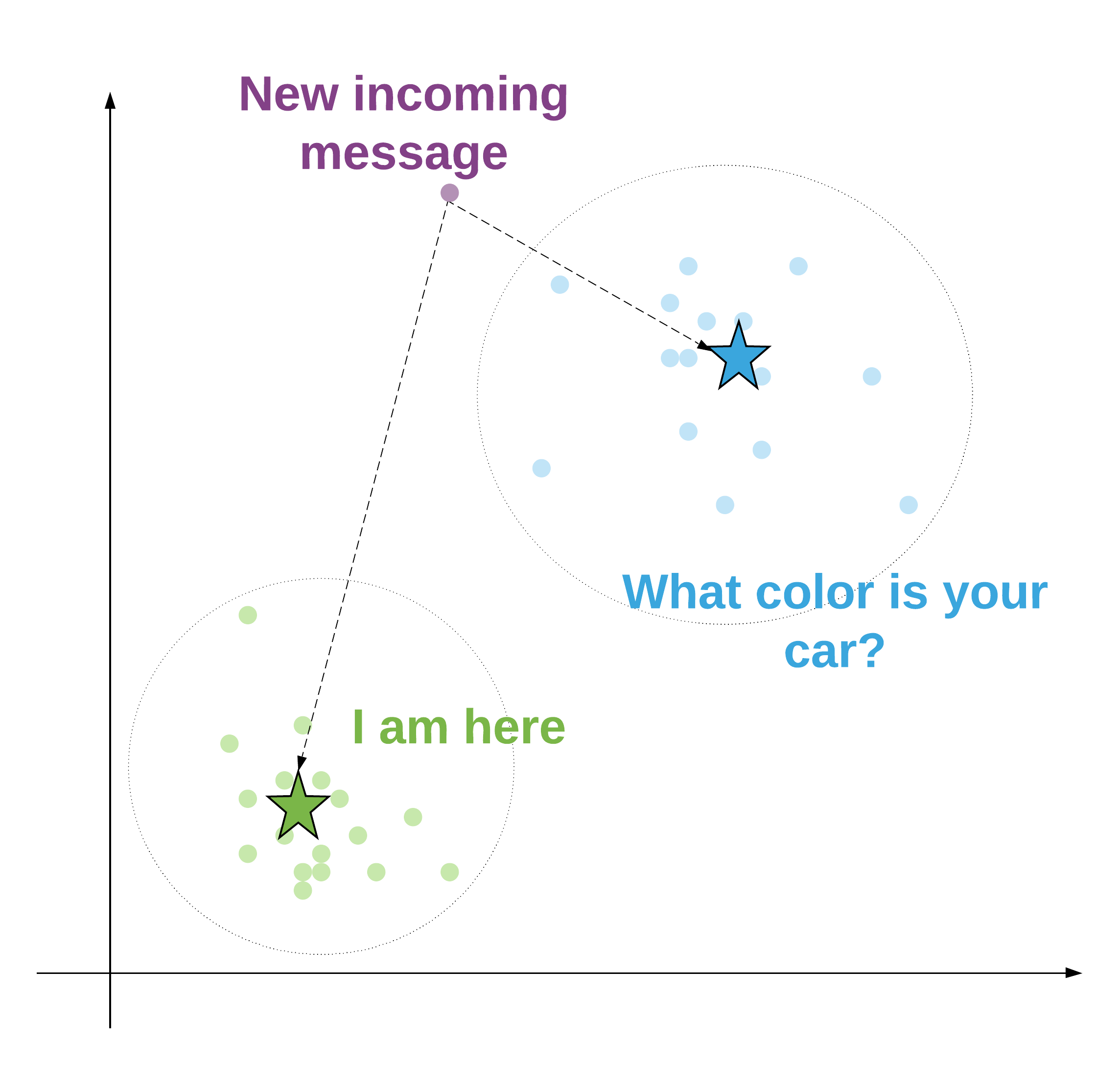}
\caption{ Illustration of the process of nearest neighbor classifier based on document embedding and cosine distance.}
\label{fig:NNC}
\end{figure}

Figure ~\ref{fig:NNC} illustrates the process of nearest-neighbor classifier using document embedding. 
First, with the trained Doc2vec model, noted as $M$, we can obtain document vector $d_j^i$ for any document, which is the dense vectors representing the $i$th document which belongs to $j$th intent class. Using labeled data, we then compute the \textbf{centroid} $D_j$ of each intent class from the dense vectors as: 

\begin{equation} \label{eq:2}
D_j= \frac{1}{N_j}\sum_{i=1}^{N_j} d_j^i
\end{equation}

\noindent where $N_j$ is the number of labeled messages of $j$th intent class.
Each intent class now is represented by this centroid. 

During the inference stage, an inference step is taken to use $M$ to compute the paragraph vector for a new paragraph $m^k=w_1, w_2, ... w_n$,
where $w_1, w_2, ... w_n$ are the word tokens. We obtain the corresponding dense vector 
$$d^k = M(m^k)$$
Using labeled data, we then compute the vector cosine distance between the message vector and each of the intents' centroids and pick top $K$ closest intents measured by cosine distance as top-$K$ predictions of intent. 

$$C_k^j = \frac{d^k \cdot D_j}{\left\|\mathbf{d}^{k}\right\|\cdot\left\|\mathbf{D}_{j}\right\|}$$


Figure~\ref{fig:NNC} illustrates a toy example with only two intent classes. An incoming message is mapped to the embedding space. As it is closer to the \textit{What color is your car?} intent centroid, it would be classified as such rather than \textit{I am here}.

\begin{figure*}[t]
\includegraphics[width=0.7\textwidth]{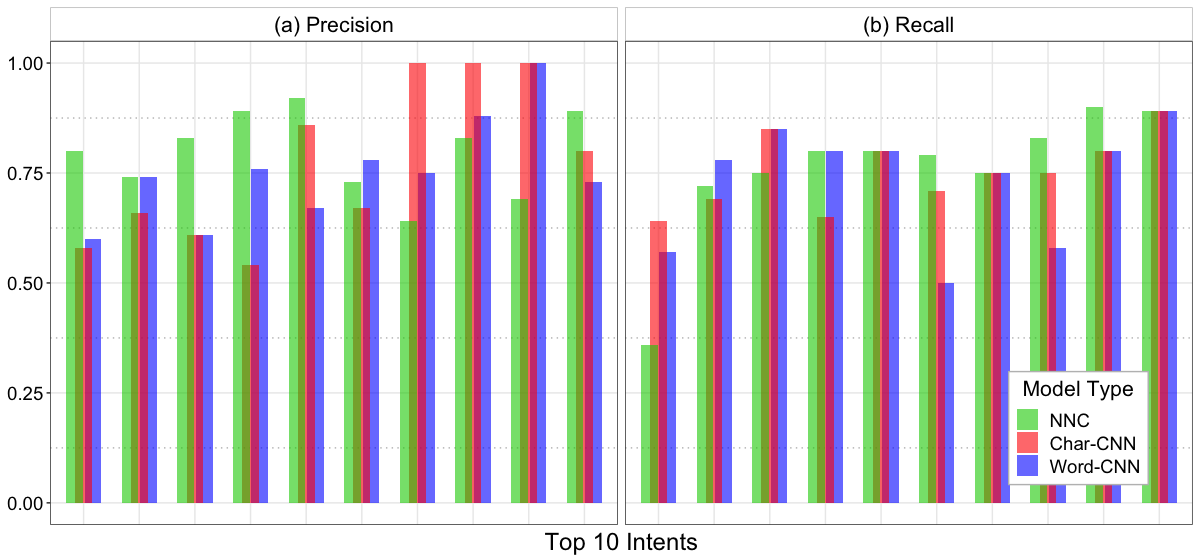}
\caption{(a) Precision and (b) Recall of the models for top-$10$ intents.}
\label{fig:prec_recall}
\end{figure*}

\subsection{Reply Retrieval}
Once the system detects the intent for the message, reply retrieval becomes relatively straightforward as the topic domain is specific to Uber's in-app communications.
For the example shown in Figure~\ref{figure:intent_reply}, the reply to a message with \textit{where are you} intent has only a small set of possible variations given its context as a response on the Uber platform. 
In this case, there are only a couple of answers such as \textit{Yes, I am omw}, \textit{Sorry, in traffic} and so on, depending on the location of the driver partner. 

Here, we leverage historical conversation pairs to find the most frequent reply candidates for each intent class. 
In essence, we perform intent classification on all messages and map out the intents of each message in a conversation. For each turn of the conversation, the intent of the \textit{incoming message} is paired with the intent of the \textit{response message}. For a particular intent of \textit{incoming message}, we measure the frequency of the intents from the \textit{response messages}, and select the most frequent ones as well as all possible variations as candidate replies.
After that, our content team performs one more round of augmentation and reordering to make the candidate replies as easily understood and accurate as possible. This whole process creates the intent-reply mapping for reply retrieval. 

During serving time, in order to gain more coverage, we pick the top $K$ predicted intents and dynamically de-duplicate repeated reply candidates by order. For instance, when we get predicted intents $I_1$ and $I_2$, we first look up the intent-reply mapping $${I_1 : R_1, R_2; I2 : R_1, R_3, R_4}.$$ Instead of providing reply $R_1$ twice, we merge them and keep their order. So the final smart reply list is $$[R_1, R_2, R_3, R_4]$$ 

In addition, corner cases such as extremely short messages (e.g., having only one word) and low confidence predictions (e.g., multi-intent messages) are handled by rules rather than our algorithm.

\begin{table}
\centering
\begin{tabular}{l|r}
Model  & Accuracy \\\hline
NNC & 0.759 \\
Word-CNN & 0.756 \\ 
Char-CNN & \textbf{0.772} \\
Frequency-based & 0.155
\end{tabular}
\caption{ Model accuracy of intent detection.}
\label{table:model_accuracy}
\end{table}

\begin{figure}
\includegraphics[width=0.4\textwidth]{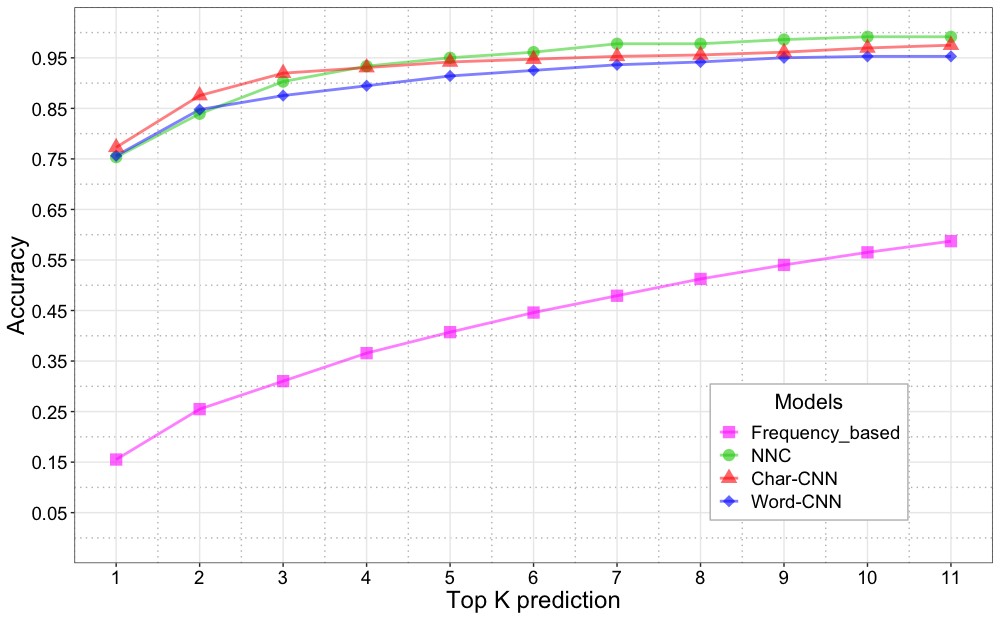}
\caption{Top-$K$ accuracy of intent detection.}
\label{fig:topK_accuracy}
\end{figure}


\section{Experiment Analysis} \label{experiment}

In this section, we evaluate the intent detection task and report overall performance for the approaches described in the above section.

\subsection{Results}
\textbf{Model Accuracy}:
One of the most important metrics for evaluating our system is the overall accuracy for intent detection as we strictly control the number of replies for each intent in the product. Table~\ref{table:model_accuracy} shows the model accuracy for the four different models we experiment with. The naive frequency-based approach has an accuracy of $15.5\%$ which is simply the population of the top-$1$ intent classes. The best performing model is Char-CNN model with an accuracy of $77.2\%$, followed by the NNC with an accuracy of $75.9\%$. Word-CNN performs slightly ($75.6\%$) worse compared to NNC.

\begin{figure}
\centering
(a)
\\
\includegraphics[width=7cm]{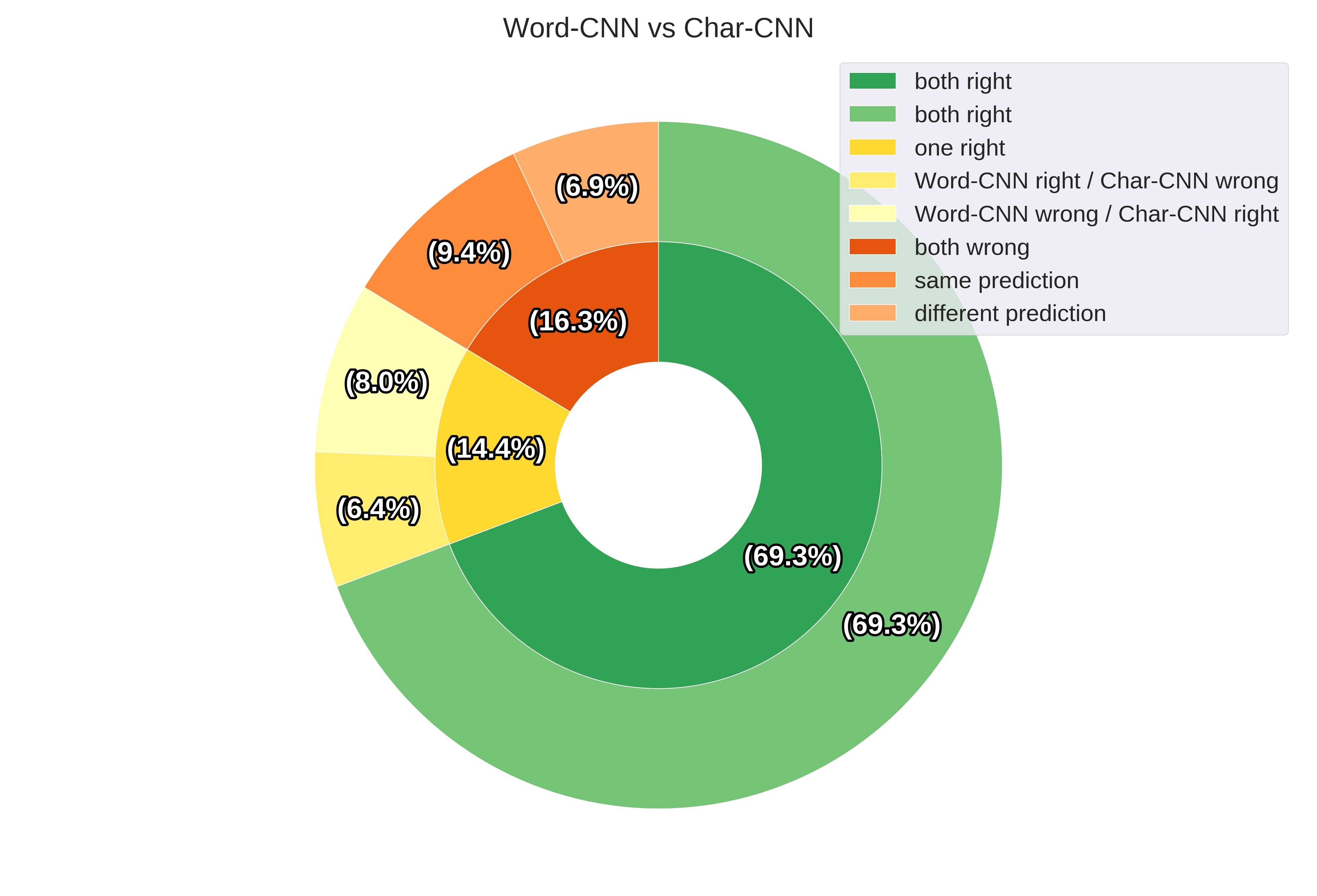}
\\
(b)
\\
\includegraphics[width=7cm]{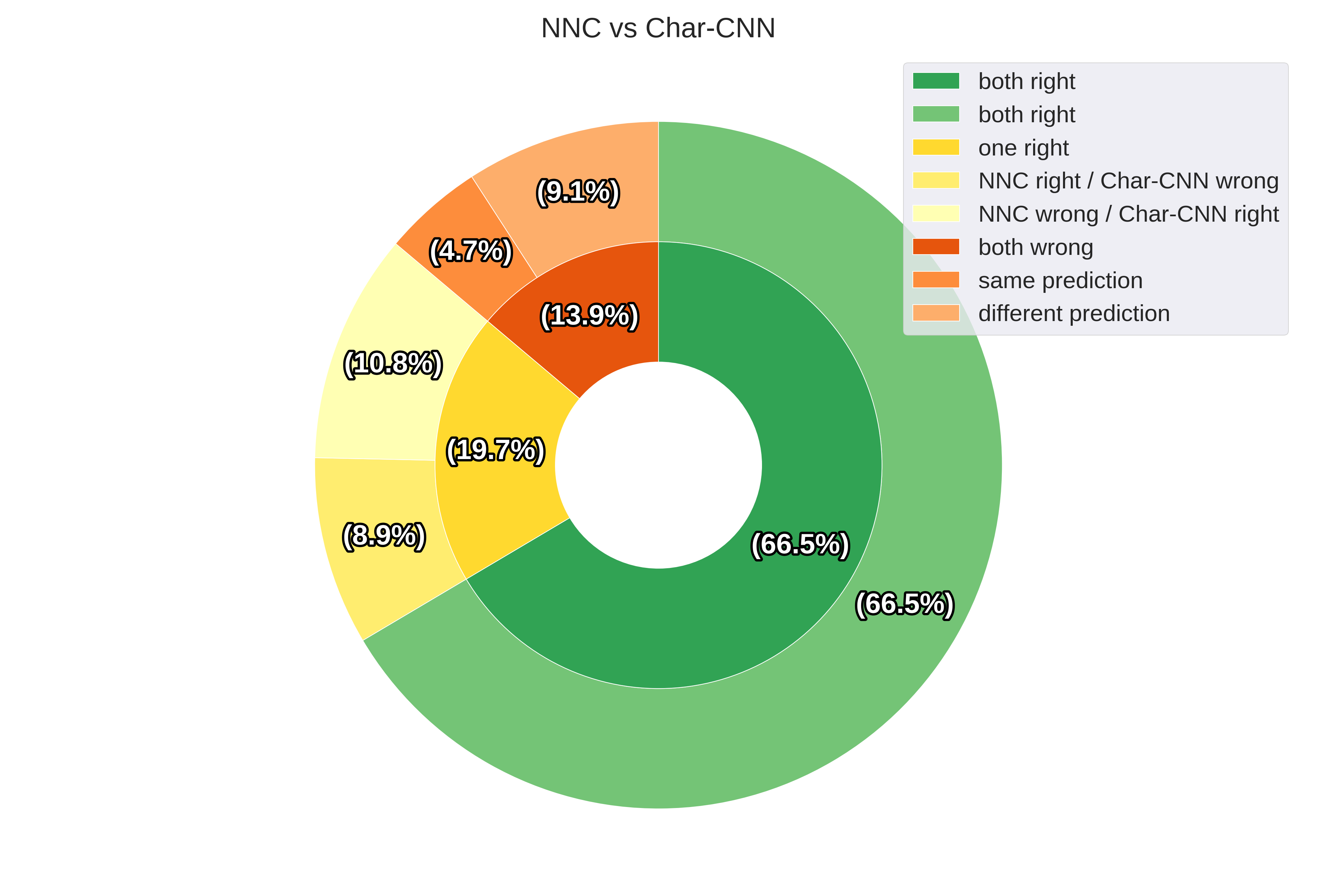}
\caption{Model prediction comparison: (a) Word-CNN vs Char-CNN models, (b) NNC vs Char-CNN models. The percentage of test samples falling into different buckets are plotted.}
\label{fig: donut_plot}
\end{figure}

Figure~\ref{fig:topK_accuracy} further shows the top-$K$ accuracy of the four approaches. The overall trend with increasing $K$ agrees with the top-1 accuracy except that NNC performs slightly better than Char-CNN after $K=5$. Except the naive frequency-based approach, all three models reach $>90\%$ accuracy at $K=4$. 
Given the small amount of labeled data (thousands) we have for training, it is not surprising that NNC performs comparable to the two deep learning architectures, as typically deep learning approaches start to be advantageous when the training data size is large enough. 

\begin{figure}
\includegraphics[width=0.4\textwidth]{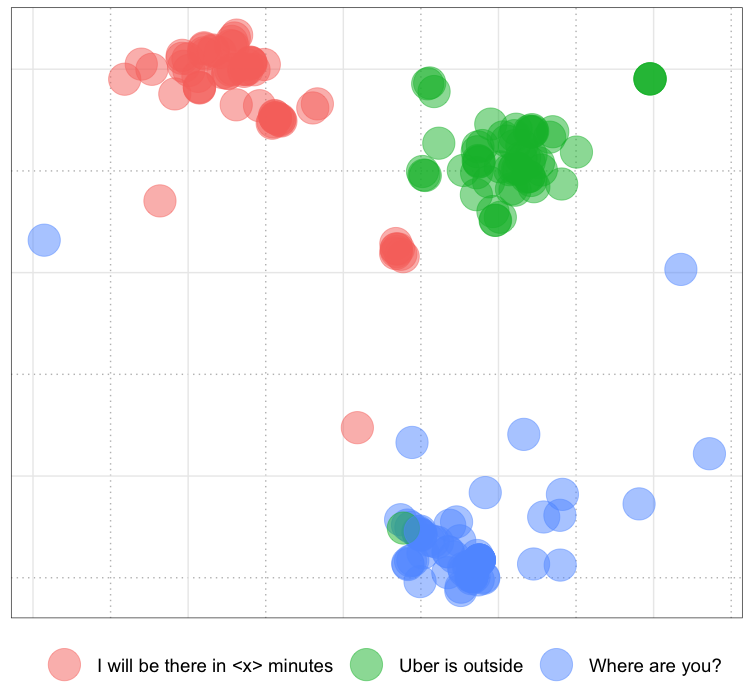}
\caption{In this two-dimensional t-SNE projection of sentence embedding, the model clusters messages around intent.}
\label{fig:message_cluster}
\end{figure}


Figure~\ref{fig:prec_recall} shows the precision and recall of NNC, Word-CNN and Char-CNN on the top-$10$ intents. 
Specifically, for precision, NNC model shows relatively even performance across all top-$10$ classes. While deep learning models (Char-CNN) in particular show lower precision for the top-$5$ intents compared to the remaining ones. This highlights a key difference between the NNC model and deep learning architectures. Because NNC is non-parametric, it is less biased towards popular classes. The pattern observed above for precision is reversed when considering recall: deep learning models are performing better than NNC for top classes (top-$3$ in particular) as shown in Figure~\ref{fig:prec_recall}(b) due to the same biased towards predicting popular classes compared to NNC.



\textbf{Model Complementary}:
Next, we look at how complementary the predictions are between NNC and deep learning models using Ludwig \cite{ludwig}. As shown in Figure~\ref{fig: donut_plot}(a), Word-CNN has a rather large overlap with Char-CNN in predictions: they have $69.3\%$ predictions being correct at the same time, and $9.4\%$ predictions being the same but wrong at the same time. Together, they made the same predictions on $78.7\%$ test samples. In contrast, Figure~\ref{fig: donut_plot}(b) shows that NNC and Char-CNN have less overlap in their predictions: $66.5\%$ being right and $4.7\%$ being the same but wrong at the same time. 
Looking at the portion of predictions where one model being right and the other being wrong, we find that NNC and Char-CNN have $19.7\%$ such predictions compared to $14.4\%$ from Word-CNN and Char-CNN.
It confirms that non-parametric model NNC is indeed more complementary to Char-CNN than Word-CNN. 

\begin{table*}
\begin{tabular}{|p{5cm}|p{4cm}|p{3cm}|p{3cm}|}
\hline
\textbf{Message content} & \textbf{First prediction} & \textbf{Second prediction} & \textbf{Label} \\ \hline
\textit{This Uber is for my daughter. She's going to school and coming right back. Thanks} & You are picking up \textless{}person\textgreater{} & I'm going to \textless{}loc\textgreater{} & I'm going to \textless{}loc\textgreater{} \\ \hline
\textit{Ok I'll drive on the Main Street} & Wrong side & I am at \textless{}loc\textgreater{} & I'm going to \textless{}loc\textgreater{} \\ \hline
\textit{I will come to 51 and 6} & Come to \textless{}loc\textgreater{} & Can we meet at \textless{}loc\textgreater{}? & I'm going to \textless{}loc\textgreater{} \\ \hline
\end{tabular}
\caption{Examples of prediction errors by NNC model on \textit{I am going to <loc>} intent.}
\label{table:Error analysis}
\end{table*}

\textbf{Analysis of NNC Model:}
Finally, in order to better understand the underlying mechanism of such a good performance for a simple nearest-neighbor classifier, we look at the embedding representation of the messages and their corresponding labels.
Figure~\ref{fig:message_cluster} shows examples of three different intent classes in a t-SNE plot. Surprisingly, the message embedding vectors are clustered for each intent even after projected down to $2d$ space. This confirms the high quality of the message embeddings and its capability to capture semantic meanings of different intents even though most messages are rather short and can contain various typos and abbreviations. The separation between the three classes is also rather pronounced. As a result, it is expected that a very simple non-parametric nearest-neighbor classifier can perform on par with sophisticated deep learning architectures such as Char-CNN.

\begin{figure}
\includegraphics[width=0.4\textwidth]{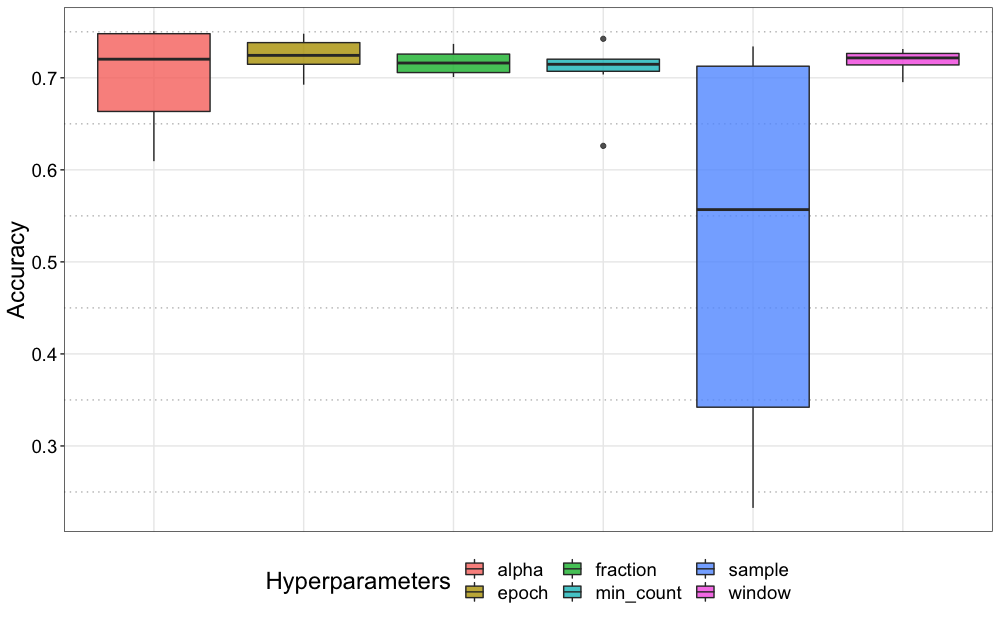}
\caption{Hyper-parameter search for Doc2vec model with distributed bag of words (DBOW). The intent detection accuracy is plotted against several parameters showing both the average and standard deviation.}
\label{fig:hyper_search}
\end{figure}

The above analysis demonstrates that the quality of document embedding is critical to the good performance of NNC. 
Furthermore, we perform a hyperparameter search to understand the correlation between the hyperparameters of the Doc2vec model and the intent detection performance using \textit{gensim} \cite{rehurek_lrec}. Empirically, we find that document vectors with distributed bag of words (DBOW) work better than those obtained with distributed memory (DM) for intent detection. Figure~\ref{fig:hyper_search} shows the hyperparameter search for DBOW and its impact on the downstream classification accuracy. It is clear that the accuracy is rather sensitive to the parameters \textit{alpha} and \textit{sample} but varies very little for all the other parameters. Specifically, \textit{alpha} is the initial learning rate of DBOW and we found that a moderate learning rate around $0.02$ is optimal. \textit{sample} determines the amount of down-sampling on high-frequency words. Empirically, we observed that a small \textit{sample} gives the best performance, implying that little or no down-sampling is necessary. It is therefore reasonable to conclude that high-frequency words are key to capture the semantic meaning of the short chat messages, and thus crucial for our intent detection task.

\textbf{Model Deployment:}
Due to its simplicity to develop and deploy, its requirement of small amount of labeled data, and its advantage of speedy inference, we decided to deploy the NNC model in production for our smart reply OCC system.
Through experimentation, we observed that $23\%$ of trips in English-speaking countries on the Uber platform involved two-way in-app communications between riders and driver-partners.
Among these in-app communications, over $71\%$ of them adopted the smart replies suggested by OCC to speedup and smooth out the process. Such an adoption rate is consistent with the system accuracy on intent detection.

\subsection{Error Analysis of NNC Model}

The NNC model has relatively high performance on most classes, as discussed above. In this section, we conduct an error analysis on intents where the NNC model doesn't perform very well to understand the weakness of the model. One such intent is \textit{I am going to <loc>}. 
Table~\ref{table:Error analysis} shows the raw message and its top $2$ predictions for this intent class in the test set. For the first message, it contains dual intents. The first part of the message informs the driver-partner that \textit{it is for another person}. The second half tells the driver-partner \textit{where she is going}, which matches the model prediction. Multi-intent issues are challenging and beyond the current design of our system.
Regarding the second and third messages, the algorithm misclassifies the intents but was able to capture the \textit{location} information correctly.
This analysis points out directions for further improvements of our system in the future.

\section{Conclusions and Future Work} \label{future_work}

In conclusion, we introduce a novel smart reply system designed to speed up in-app communication between riders and driver-partners on Uber's platform.
Our smart reply system is unique in handling short chat messages with various non-canonical text data.
The algorithm we adopt also has the advantage of requiring a rather small amount of labeled data to achieve a relatively high performance.
In contrast to existing smart reply systems, we break the task down into two steps of intent detection and reply retrieval instead of directly predicting reply.
For the task of intent detection, we experimented with four models, and showed that a simple approach of nearest-neighbor classifier together with document embedding proves to be powerful enough to achieve an accuracy of $\geq75\%$, which is comparable with the two deep learning models. Further analysis reveals that the non-parametric NNC model is more complementary to Char-CNN than Word-CNN, as Char-CNN and Word-CNN belong to the same class of deep learning architecture. Finally, we analyze the document embeddings and uncover that the key to the success of NNC model is its high quality embeddings which provides clear separations between different intent classes. Due to the advantages of easy development and deployment, and the fast inference, the NNC model is deployed in production to serve the traffic of Uber's smart reply system. Through experimentation, we observed that $\geq71\%$ of all two-way communications in English on Uber's in-app messaging platform adopted the smart replies recommended by our OCC system to speed up the communication process.

In the future, there are several areas where we can further improve the system. 
First, the current system only uses the message itself as a feature for intent detection.
Including additional features around the trip can certainly provide better intent modeling.
Second, the reply retrieval is done using static mapping from intent to replies.
Dynamic reply retrieval holds great promise for providing more context-aware replies.
This can be achieved by ranking the replies dynamically using a ranking algorithm by taking into account the contextual information.
Lastly, active learning feedback loop can be another avenue to steadily correct the errors made by the models and improve the system performance.

\section{Acknowledgments}
The authors wish to thank Uber's Conversational AI, Applied Machine Learning, Communications Platform, and Michelangelo teams, Anwaya Aras, Chandrashekar Vijayarenu, Bhavya Agarwal, Lingyi Zhu, Runze Wang, Kailiang Chen, Han Lee, Molly Vorwerck, Jerry Yu, Monica Wang, Manisha Mundhe, Shui Hu, Zhao Zhang, Hugh Williams, Lucy Dana, Summer Xia, Tito Goldstein, Ann Hussey, Yizzy Wu, Arjun Vora, Srinivas Vadrevu, Huadong Wang, Karan Singh, Arun Israel, Arthur Henry, Kate Zhang, and Jai Ranganathan.








\bibliographystyle{ACM-Reference-Format}
\balance
\bibliography{bibliography}

\end{document}